\documentclass[10pt, preprint]{article} 
\usepackage{tmlr}

\usepackage{hyperref}
\usepackage{url}

\usepackage{booktabs}
\usepackage{amsfonts}
\usepackage{nicefrac}
\usepackage{microtype}
\usepackage{xcolor}

\usepackage{graphicx}
\usepackage{subcaption}
\usepackage{caption}
\usepackage{enumitem}
\usepackage{amsmath}
\usepackage{multirow}
\usepackage{multicol}
\usepackage{algorithm}
\usepackage{algpseudocode}
\usepackage{amsthm}
\theoremstyle{definition}
\newtheorem{definition}{Definition}[section]

\title{Enhancing One-run Privacy Auditing with Quantile Regression-Based Membership Inference}

\newcommand*\samethanks[1][\value{footnote}]{\footnotemark[#1]}

\author{\name Terrance Liu
        \email terrancl@cmu.edu
        \\ \addr Carnegie Mellon University
        \AND
        \name Matteo Boglioni\thanks{Order alphabetically by last name.}
        \email mboglioni@ethz.ch
        \\ \addr ETH Zurich
        \AND
        \name Yiwei Fu\samethanks
        \email yiweif@cmu.edu
        \\ \addr Carnegie Mellon University
        \AND
        \name Shengyuan Hu\samethanks
        \email shengyuanhu@cmu.edu
        \\ \addr Carnegie Mellon University
        \AND
        \name Pratiksha Thaker\samethanks
        \email pthaker@cmu.edu
        \\ \addr Carnegie Mellon University
        \AND
        \name Zhiwei Steven Wu\samethanks
        \email zstevenwu@cmu.edu
        \\ \addr Carnegie Mellon University
  }

\def\month{MM}  
\def\year{YYYY} 
\def\openreview{\url{https://openreview.net/forum?id=XXXX}} 

\newcommand{\eps}{\varepsilon}

\begin{document}

\maketitle

\begin{abstract}

Differential privacy (DP) auditing aims to provide empirical lower bounds on the privacy guarantees of DP mechanisms like DP-SGD. While some existing techniques require many training runs that are prohibitively costly, recent work introduces one-run auditing approaches that effectively audit DP-SGD in white-box settings while still being computationally efficient. However, in the more practical black-box setting where gradients cannot be manipulated during training and only the last model iterate is observed, prior work shows that there is still a large gap between the empirical lower bounds and theoretical upper bounds. Consequently, in this work, we study how incorporating approaches for stronger membership inference attacks (MIA) can improve one-run auditing in the black-box setting. Evaluating on image classification models trained on CIFAR-10 with DP-SGD, we demonstrate that our proposed approach, which utilizes quantile regression for MIA, achieves tighter bounds while \textit{crucially} maintaining the computational efficiency of one-run methods.

\end{abstract}

\section{Introduction}

Differential privacy (DP) \citep{dwork2006calibrating} has become an effective, practical framework for 
specifying and ensuring privacy guarantees of statistical algorithms,
including stochastic gradient descent (DP-SGD) for training large models privately \citep{chaudhuri2011differentially, abadi2016deep}.
While DP provides an upper bound on the privacy guarantee $\eps$ of the algorithm, 
it is useful to additionally have a \emph{lower bound} on $\varepsilon$ to validate it in practice and potentially detect errors in implementations~\citep{ding2018detecting, jagielski2020auditing, tramer2022debuggingdifferentialprivacycase}.
This lower bound is derived empirically through \emph{privacy auditing}.

DP Auditing often requires training a model hundreds---if not thousands---of times, inducing heavy computational requirements that simply don't scale when auditing larger models \citep{tramer2022debuggingdifferentialprivacycase}. These costs are further exacerbated by the computational costs of calculating per-example gradients in DP-SGD. Despite recent advancements in computational efficiency \citep{nasr2023tight}, multiple-run auditing still incurs overheads that can lead to prohibitively costly experiments \citep{annamalai2024nearlytightblackboxauditing}. In light of these problems, \citet{steinke2023privacy} introduce a new framework requiring only a single run. Framed as a guessing game, the goal is to identify among a set of ``canary'' examples the ones that were seen during training. If one is able to make more guesses correctly, then one can establish higher empirical lower bounds on $\varepsilon$.

We view these types of guessing games for DP auditing as a form of membership inference \citep{shokri2017membership}, where the goal is determine if a given sample was used in training a machine learning model. However, \citet{steinke2023privacy} and \citet{mahloujifar2024auditing} introduce and evaluate their auditing schemes using only the simplest strategy for MIA, which can be summarized as looking at some score function (i.e., loss of the canary) and sorting (i.e., predicting that it was used in training if the loss is small and vice versa). We posit, however, that in applying this naive strategy, these auditing procedures may underestimate the empirical lower bounds for DP-SGD.

\paragraph{Contributions.} In this work, we evaluate to what extent using strong MIA methods for privacy auditing in the one-run setting can tighten empirical privacy estimates. Given that the purpose of such one-run auditing procedures is to assess privacy mechanisms while maintaining efficiency, we specifically adopt approaches for MIA introduced in \citet{bertran2023scalablemembershipinferenceattacks}, who introduce a class of attacks that compete with state-of-the-art shadow model approaches for MIA \citep{shokri2017membership, carlini2022membershipinferenceattacksprinciples} while being computationally efficient (i.e., also require one training run). 

We consider the black-box setting for auditing, where the auditor can only access the model at the final training step.
Evaluating on image classification models trained on CIFAR-10 using DP-SGD, we demonstrate that MIA significantly improves empirical lower bounds estimated from one-run procedures introduced by \citet{steinke2023privacy} and \citet{mahloujifar2024auditing}. Furthermore, we find that the advantage holds across a wide range of data settings (i.e., the number of training examples and proportion of canaries inserted into training), improving the lower bound by up to 3x in some cases.

\subsection{Additional Related Works}

In addition to those mentioned above, there have many other works that have recently studied private auditing under various scenarios. For example, rather than auditing models that are made private during training, \citet{chadha2024auditing} audit methods that are made private during inference.
\citet{pillutla2023unleashing}, on the other hand, introduce the definition of Lifted Differential Privacy (LiDP) and propose a multi-run auditing procedure that can utilize multiple, randomized canaries (similar to our one-run auditing setting, in which the auditor also inserts many canaries).
Furthermore, a variety of works have recently studied private auditing specifically under the constraints of black-box model access.
\citet{steinke2024last} study black-box auditing of DP-SGD for models with linear structure, proposing a heuristic that predicts the outcome of an audit performed on only the last training iterate.
\citet{annamalai2024nearlytightblackboxauditing} show that empirical lower bounds for black-box auditing are much tighter when models are initialized to worst-case parameters, and lastly, \citet{cebere2024tighter} study black-box auditing in the case where an adversary can inject sequences of gradients that are crafted ahead of training.

\section{Preliminaries}

At a high level, differential privacy provides a mathematical guarantee that the output distribution of an algorithm is not heavily influenced by any single data point. Formally, it is defined as the following:
\begin{definition}[Differential Privacy (DP) \citep{dwork2006calibrating}]\label{def:dp}
A randomized algorithm $\mathcal{M}:\mathcal{X}^N\rightarrow \mathbb{R}$ satisfies $(\varepsilon, \delta)$-differential privacy if 
for all neighboring datasets $D, D'$ and for all outcomes $S\subseteq \mathbb{R}$ we have
\begin{align*}
    P(\mathcal{M}(D) \in S) \leq e^{\varepsilon} P(\mathcal{M}(D') \in S ) + \delta
\end{align*}
\end{definition}

To train deep learning models with privacy guarantees, \citet{abadi2016deep} propose a differentially private form of stochastic gradient descent (DP-SGD), in which the algorithm noises and clips gradients before every update step.

\subsection{One-run auditing}

To audit models trained using DP-SGD, we consider the following ``one-run'' auditing procedures:
\begin{enumerate}[itemsep=1pt, leftmargin=20pt]
    \item \textbf{\citet{steinke2023privacy}.} \citet{steinke2023privacy} first developed the notion of auditing in one training run. Rather than training many models on neighboring datasets that differ on \textit{single} examples, their auditing scheme requires training only a single model on a dataset with \textit{many} ``canary'' examples. Specifically, these canaries are randomly sampled from a larger set of canaries. The auditor then attempts to predict which canaries were in and not in the training set (with abstentions are allowed). The final empirical lower bound is determined by how many guesses were made and how many were correct.
    
    \item\textbf{\citet{mahloujifar2024auditing}.} More recently, \citet{mahloujifar2024auditing} present an alternative approach, which they show provides better privacy estimates in the one-run setting. Rather than having a single set of canaries, \citet{mahloujifar2024auditing}'s method first constructs a set of canary sets of size $K$, where a random example in each canary set is using in training. Here, the goal is to guess which of the $K$ canaries in each set was used in training. As in \citet{steinke2023privacy}, abstentions are also allowed, and again, the empirical lower bound is determined by the number of guesses made and the number that are correct. 
\end{enumerate}
We present in Algorithms \ref{alg:auditor_one_run} and \ref{alg:reconstruction_one_run} the auditing procedures for \citet{steinke2023privacy} and \citet{mahloujifar2024auditing}, respectively.

We note that in Algorithms \ref{alg:auditor_one_run} and \ref{alg:reconstruction_one_run}, we make minor changes to the notation compared to how they were original introduced in their respective works \citep{steinke2023privacy, mahloujifar2024auditing}. In this way, we make the notation of the two algorithms consistent with each other. For example, we now let $n$ denote the total number of examples used in training (rather than the total number of auditing and non-auditing examples in \citet{steinke2023privacy}) and $m$ be the total number of canaries (rather than canary sets in \citet{mahloujifar2024auditing}). In Algorithm \ref{alg:auditor_one_run}, exactly half of the canaries are randomly sampled such that the data partitioning is exactly equivalent to \citet{mahloujifar2024auditing} when the canary set size is $K=2$.

\begin{algorithm}[t!]
\caption{Auditor with One Training Run}\label{alg:auditor_one_run}
\begin{algorithmic}[1]
    \Require Algorithm to audit $\mathcal{A}$, target number of examples to train on $n$, scoring function \textsc{Score}
    \Require Number of positive and negative guesses $k_+$ and $k_-$ respectively, 
    \Require $x \in \mathcal{X}^{m + r}$ consisting of $m$ auditing examples (a.k.a. canaries) and $r$ non-auditing examples, where $n = r + \frac{m}{2}$
    \State Randomly assign $S_i = +1$ to half of the $m$ canaries and $S_i=-1$ to the other half. Set $S_i = 1$ for all remaining examples $i \in [m + r] \setminus [m]$.
    \State Partition $x$ into $x_{\text{IN}} \in \mathcal{X}^{n_{\text{IN}}}$ and $x_{\text{OUT}} \in \mathcal{X}^{n_{\text{OUT}}}$ according to $S$, where $n_{\text{IN}} + n_{\text{OUT}} = n$. Namely, if $S_i = 1$, then $x_i$ is in $x_{\text{IN}}$; and, if $S_i = -1$, then $x_i$ is in $x_{\text{OUT}}$.
    \State Run $\mathcal{A}$ on input $x_{\text{IN}}$ with appropriate parameters, outputting $w$.
    \State Compute the vector of scores $Y = (\textsc{Score}(x_i, w) : i \in [m]) \in \mathbb{R}^m$
    \State (i.e., $T \in \{-1, 0, +1\}^m$ maximizes $\sum_{i} T_i \cdot Y_i$ subject to $\sum_{i} |T_i| = k_+ + k_-$ and $\sum_{i} T_i = k_+ - k_-$).
    \State \Return The vector $S \in \{-1, +1\}^m$ indicating the true selection and the guesses $T \in \{-1, 0, +1\}^m$.
\end{algorithmic}
\end{algorithm}

\begin{algorithm}[t!]
\caption{Reconstruction in one run game}\label{alg:reconstruction_one_run}
\begin{algorithmic}[1]
    \Require Algorithm to audit $\mathcal{A}$, target number of examples to train on $n$, scoring function \textsc{Score}
    \Require Number of guesses $k$ respectively
    \Require $M = \frac{m}{K}$ sets (of size $K$) of canary examples and $r$ non-auditing examples, where $n = r + \frac{m}{K}$ (assume that $m$ is a multiple of $K$).
    \State Let $\mathcal{C} = \{x_j^i \}_{i \in [M], j \in [K]}$ be the matrix of canaries 
    \State Let $u = (u_1, \dots, u_M)$ be a vector uniformly sampled from $[K]^M$.
    \State Let $S = \{x_{u_i}^i : i \in [M]\}$.
    \State Run mechanism $\mathcal{A}$ on $S \cup \mathcal{T}$ to get output $w$.
    \State Compute the matrix of scores $Y = (\textsc{Score}(x_j^i, w): i \in [M], j \in [K]) \in \mathbb{R}^{M \times K}$
    \State Use scores $Y$ to run a reconstruction attack on $w$ to obtain a vector $v = (v_1, \dots, v_M) \in ([K] \cup \{\bot\})^M$ in which the number of guesses is $k$ (i.e., $k = \sum_i^M \mathbf{1}\{ v_i \ne \bot \})$
    \State (i.e., $v$ is a vector guessing the index of the canary in each set that was used in training. $\bot$ indicates an abstention.)
    \State \Return The vector $v$
\end{algorithmic}
\end{algorithm}

\paragraph{Black-box auditing.}

\citet{nasr2023tight} presents two main threat models:
\begin{itemize}[itemsep=1pt, leftmargin=20pt]
    \item White-box access: The auditor has full access throughout the training process to both model's weights and gradients, being able to inject arbitrarily-designed gradients at each update step
    \item Black-box access (with input space canaries): This approach is more restrictive. The auditor is only able to insert training samples in the dataset and observe the model at the end of training.
\end{itemize}

In our work, we study the \textit{black-box} setting that does not allow modifications to the training procedure (i.e., modifying gradients like in white-box setting with Dirac gradients \citep{nasr2023tight, steinke2023privacy, mahloujifar2024auditing} or in an alternative black-box setting studied in \citet{cebere2024tighter} that allows gradient sequences to be inserted). This threat model is often more practically relevant and includes settings such as publishing the final weights of an open-sourced model. As shown in \citet{nasr2023tight} and \citet{steinke2023privacy}, the gap between the empirical lower bound and theoretical upper bound is generally still large in the black-box setting, suggesting that this area of research may still be underexplored.\footnote{\citet{mahloujifar2024auditing}, for example, do not evaluate their proposed method in the black-box setting at all.} 


\section{Applying (Efficient) MIA to Privacy Auditing}

\begin{figure}[t!]
  \centering
  \begin{subfigure}[b]{0.48\textwidth}
    \centering
    \includegraphics[width=0.8\linewidth]{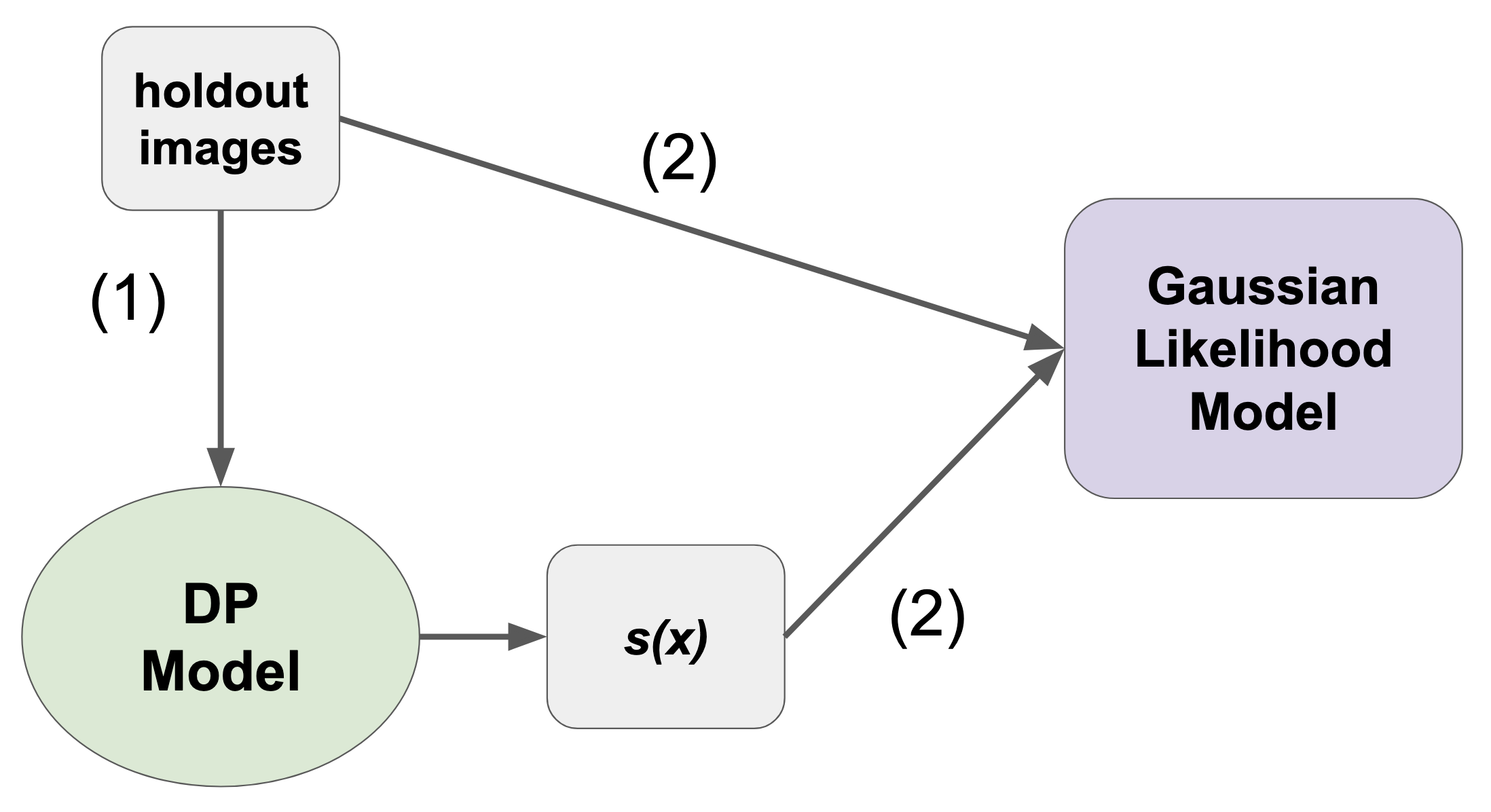}
    \caption{Train ``MIA'' regressor}
    \label{fig:quantile_a}
  \end{subfigure}
  \hfill
  \begin{subfigure}[b]{0.48\textwidth}
    \centering
    \includegraphics[width=\linewidth]{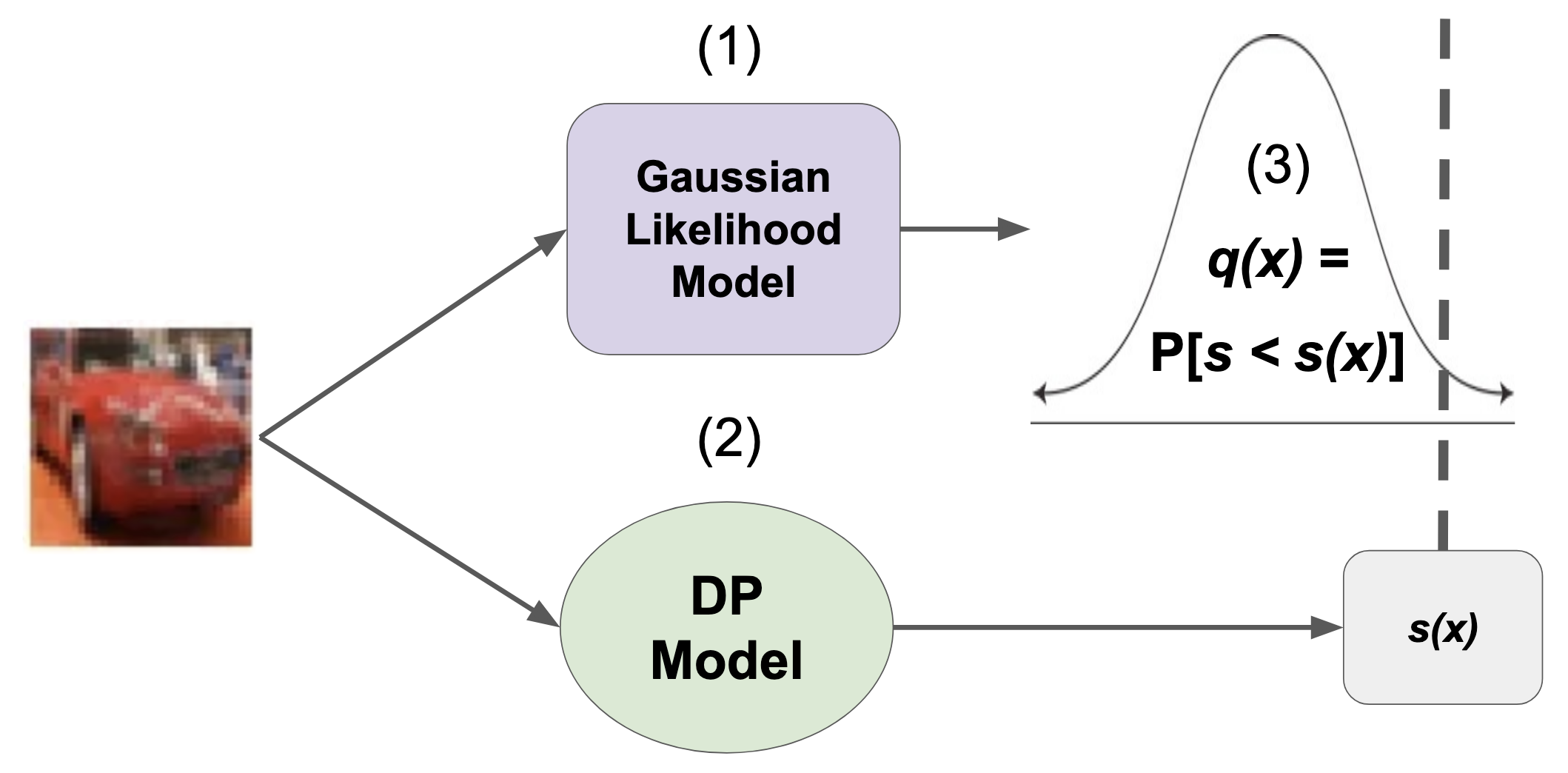}
    \caption{``Rescore'' canaries}
    \label{fig:quantile_b}
  \end{subfigure}
  \caption{
  We provide a high-level diagram describing our quantile-regression based MIA approach to auditing. In the first part \textbf{(a)}, we \textbf{(1)} calculate the score $s(x)$ (e.g., loss) using the privately trained model and \textbf{(2)}, train the Gaussian likelihood model on the images $x$ themselves and $s(x)$. Once the Gaussian likelihood model has been trained on the holdout set, in part \textbf{(b)}, we take each canary and \textbf{(1)} use the Gaussian likelihood model to output the parameters of a Gaussian distribution (i.e., $\mu$, $\sigma$). \textbf{(2)} Next, we again feed the canary into the private model to obtain $s(x)$. \textbf{(3)} Finally, we calculate our new score, $q(x) = P(s < s(x) \mid \mu, \sigma)$.
  }
  \label{fig:quantile}
\end{figure}

Membership inference often requires the attacker to train several shadow models on a random subsets of data \citep{carlini2022membershipinferenceattacksprinciples}. This approach, while effective, requires high computational demands that do not align with the goals of one-run auditing. In contrast, \citet{bertran2023scalablemembershipinferenceattacks} introduce a new class of MIA methods that relies on training a single quantile regressor on holdout data only. In doing so, they predict a sample-specific threshold for determining membership that outperforms marginal thresholds, which are equivalent to the sort and rank (by loss) procedure employed in \citet{steinke2023privacy} and \citet{mahloujifar2024auditing}.

Formally, the quantile regressor can be written as the following:
\begin{definition}[Quantile Regressor]
Given a target false positive rate $\alpha$, a quantile regressor is a model $q : \mathcal{X} \rightarrow \mathbb{R}$ trained on an holdout dataset $\mathcal{P}$ to predict the $(1-\alpha)$-quantile for the score distribution associated to each given sample:
\[\forall (x,s) \in \mathcal{P}\textrm{,} \hspace{5pt} Pr[y\leq q(x) ] = 1-\alpha\]
\end{definition}

Given the relatively small sample size in image datasets like CIFAR-10, \citet{bertran2023scalablemembershipinferenceattacks} propose an alternative method for outputting quantile thresholds in which they train a model that instead predicts the mean $\mu(x)$ and the standard deviation $\sigma(x)$ of the score $s(x)$ (e.g., loss of the model to be attacked) associated with each example $x$. The per-example threshold is then calculated based on this normal distribution (i.e., $P(s < s(x) \mid \mu, \sigma)$).

The loss can then be written as the following:
\begin{definition}[Negative Log-Likelihood for Gaussian Distributions]
The negative log-likelihood loss for a Gaussian distribution with mean \(\mu\) and standard deviation \(\sigma\) is given by:

\[
\mathcal{L}_{\text{NLL}} = \mathbb{E}_{x \sim p(x)} \left[ \frac{(x - \mu)^2}{2\sigma^2} + \log \sigma \right]
\]

where \(x \sim p(x)\) represents samples from some underlying data distribution (e.g., losses from an image classification model).
\end{definition}

In our proposed method (See Figure \ref{fig:quantile}), we also adapt this approach and train a neural network to output a Gaussian distribution for each canary image. However, rather than using as a threshold the $q$-quantile for some predetermined value of $q$ \citep{bertran2023scalablemembershipinferenceattacks, tang2024membership}, we calculate $q$ directly (i.e., the CDF $P(s < s(x) \mid \mu, \sigma)$). We then use $q$ as the input \textsc{Score} function for Algorithms \ref{alg:auditor_one_run} and \ref{alg:reconstruction_one_run}.



\section{Empirical Evaluation}

\paragraph{Auditing setup.}

For our empirical evaluation, we follow the experimental set up in prior work \citep{nasr2023tight, steinke2023privacy, mahloujifar2024auditing} and train Wide ResNet models \citep{zagoruyko2016wide} using DP-SGD on the CIFAR-10 dataset \citep{krizhevsky2009learning}. All models are trained using code provided by \citet{jax-privacy2022github}, which implements training of state-of-the-art DP CIFAR-10 models presented in \citet{de2022unlocking}. While \citet{steinke2023privacy} experimented with black-box canaries with both flipped and unperturbed class labels, we found early on that flipping labels did not improve the lower bound. Thus, given that perturbing the labels can only hurt the final DP model's accuracy, we do not flip the canary labels in our experiments.

\paragraph{Quantile regressor.}

Following \citet{bertran2023scalablemembershipinferenceattacks}, we use ConvNeXt \citep{liu2022convnet} as our model architecture for the quantile regressor. Similarly, we use as our score function the difference in logit of the true class label and the sum of the remaining logits. As shown in \citet{carlini2022membershipinferenceattacksprinciples}, this score function---in contrast to cross-entropy loss---follows a normal distribution, making it a natural choice for our approach.

\paragraph{Choice of number of guesses $k$.}

In general, empirical lower bounds on $\varepsilon$ can be quite sensitive to the number guesses made \citep{mahloujifar2024auditing}. However, it is unclear from both \citet{steinke2023privacy} and \citet{mahloujifar2024auditing} how the number of guesses was chosen to produce their main results. For example, \citet{steinke2023privacy} state that they ''evaluate different values of $k_+$ and $k_-$ and only report the highest auditing results,'' but do not specify what exact values were tested. We reached out to the authors, who told us that some values between 10 and 1000 were chosen (but not exactly how many values of $k$ were tested). Consequently, we evaluate all methods in our experiments from $10$ to the maximum number of guesses possible in multiples of $10$, and like prior work \citep{nasr2023tight, steinke2023privacy, mahloujifar2024auditing}, report the highest auditing results for each run.
\section{Results}

\begin{table}[t!]
\caption{
    We present the empirical lower bounds estimated using baseline method and quantile regression (\textit{ours}). $\varepsilon_{\textrm{or}}$ corresponds to \citet{steinke2023privacy}, $\varepsilon_{\textrm{or-fdp}}$ corresponds to \citet{mahloujifar2024auditing}, and $\varepsilon_{\textrm{or-max}}$ corresponds to max of the two. We calculate $\varepsilon$ for 5 different runs and report the average.
}
\centering
\begin{tabular}{ll|ccc}
\toprule
\multirow{2}{*}{$n$} & \multirow{2}{*}{method} & \multicolumn{3}{c}{$r=45000$, $m=5000$} \\
& & $\varepsilon_{\textrm{or}}$ & $\varepsilon_{\textrm{or-fdp}}$ & $\varepsilon_{\textrm{max}}$ \\
\toprule
\multirow{2}{*}{47500}
& baseline & 0.159 & \textbf{0.147} & 0.208 \\
& \textit{ours} & \textbf{0.210} & 0.134 & \textbf{0.253} \\
\bottomrule
\end{tabular}
\label{tab:results_50k}
\end{table}
\begin{table}[t!]
\caption{
    We present the empirical lower bounds estimated using baseline method and quantile regression (\textit{ours}) for various data settings, including when the canaries make up all ($r=0$) and half ($r=\frac{n}{2}$ of the training examples. $\varepsilon_{\textrm{or}}$ corresponds to \citet{steinke2023privacy}, $\varepsilon_{\textrm{or-fdp}}$ corresponds to \citet{mahloujifar2024auditing}, and $\varepsilon_{\textrm{or-max}}$ corresponds to max of the two. We calculate $\varepsilon$ for 5 different runs and report the average. 
}
\centering

\begin{tabular}{ll|ccc|ccc}
\toprule
\multirow{2}{*}{$n$} & \multirow{2}{*}{method} & \multicolumn{3}{c}{$r=0$, $m=2n$} & \multicolumn{3}{|c}{$r=\frac{n}{2}$, $m=n$} \\
& & 
$\varepsilon_{\textrm{or}}$ & $\varepsilon_{\textrm{or-fdp}}$ & $\varepsilon_{\textrm{max}}$ &
$\varepsilon_{\textrm{or}}$ & $\varepsilon_{\textrm{or-fdp}}$ & $\varepsilon_{\textrm{max}}$ \\
\toprule
\multirow{2}{*}{5000} 
    & baseline & 0.181 & 0.175 & 0.237 & \textbf{0.299} & 0.234 & 0.393 \\
    & \textit{ours} & \textbf{0.280} & \textbf{0.240} & \textbf{0.364} & 0.279 & \textbf{0.486} & \textbf{0.503} \\
\midrule
\multirow{2}{*}{10000}
    & baseline & \textbf{0.202} & 0.172 & 0.216 & 0.227 & 0.115 & 0.241 \\
    & \textit{ours} & 0.201 & \textbf{0.339} & \textbf{0.364} & \textbf{0.341} & \textbf{0.217} & \textbf{0.356} \\
\midrule
\multirow{2}{*}{20000}
    & baseline & 0.055 & 0.086 & 0.098 & 0.128 & 0.191 & 0.204 \\
    & \textit{ours} & \textbf{0.146} & \textbf{0.246} & \textbf{0.268} & \textbf{0.165} & \textbf{0.313} & \textbf{0.324} \\
\bottomrule
\end{tabular}
\label{tab:results_vary_n}
\end{table}

All results reported in Tables \ref{tab:results_50k} and \ref{tab:results_vary_n} are averages over the maximum lower bound (with 95\% confidence) over 5 different runs, each of which is conducted on a different random sample of the dataset. In these tables, $\varepsilon_{\textrm{or}}$ corresponds to \citet{steinke2023privacy} and $\varepsilon_{\textrm{or-fdp}}$ corresponds to \citet{mahloujifar2024auditing}. In addition, we consider the setting in which one considers the choice of auditing procedure (i.e.,  \citet{steinke2023privacy} vs \citet{mahloujifar2024auditing}) as an additional parameter that can be chosen by the auditor.\footnote{Similarly to how \citet{steinke2023privacy} report the maximum over lower bounds produced by flipping and not flipping labels.}
In this case, we take the max of $\varepsilon_{\textrm{or}}$ and $\varepsilon_{\textrm{or-fdp}}$ for each run, which we denote as $\varepsilon_{\textrm{max}}$, and again report the average over 5 runs in Tables \ref{tab:results_50k} and \ref{tab:results_vary_n}.

In Table \ref{tab:results_50k}, we present our results when auditing a model trained with $n=47500$ examples where $m=5000$ and $r=45000$\footnote{We note that this data setup corresponds to the experiments described in \citet{steinke2023privacy} under their notation of $n=50000$ and $m=5000$. While both \citet{steinke2023privacy} and \citet{mahloujifar2024auditing} audit this model in the white-box setting, neither report results for it in the black-box setting.}. For our method, we use the remaining $10000$ holdout set examples to train the quantile regression model.
In Table \ref{tab:results_vary_n}, we run experiments similar to those found in \citet{steinke2023privacy} for the black-box setting, where the number of training examples $n$ is smaller. For each choice of $n$, we run experiments for both when $r=0$ (all training examples are canaries) and $r= \frac{n}{2}$ (half of the training examples are canaries). In these experiments, we randomly sample $20000$ examples out of the remaining holdout set examples to train our quantile regression model. 

In most cases, we find that our method achieves higher auditing results, regardless of data setting (i.e., choices of $n$, $m$, and $r$) and all auditing procedures ($\varepsilon_{\textrm{or}}$, $\varepsilon_{\textrm{or-fdp}}$, and $\varepsilon_{\textrm{max}}$) In cases where the baseline performs better, the difference between it and our method is small (e.g., difference of $0.20$ for $n=5000$, $r=\frac{n}{2}$). Our results strongly indicate that better member inference attacks can improve DP-SGD auditing and suggest that in general, MIA methods should be incorporated into auditing experiments when applicable.

\paragraph{Additional insights.}
Lastly, we present additional observations about how one-run auditing procedures operate in the black-box setting. First, we note that generally speaking, we observe no clear winner between \citet{steinke2023privacy} and \citet{mahloujifar2024auditing} in the black-box setting, in contrast to the white-box setting in which \citet{mahloujifar2024auditing} achieves much tighter auditing results compared to \citet{steinke2023privacy}. In all cases, the average $\varepsilon_{\textrm{max}}$ strictly dominates both $\varepsilon_{\textrm{or}}$ and $\varepsilon_{\textrm{or-fdp}}$, further suggesting that one auditing procedure does not consistently outperform the other. In addition, while \citet{steinke2023privacy} posit that when all training examples are canaries ($r=0$), one can achieve higher auditing results, Table \ref{tab:results_vary_n} does not clearly corroborate this hypothesis (if anything, the auditing procedures estimate slightly higher lower bounds when $r=\frac{n}{2}$). We leave further investigation of such observations to future work.
\section{Conclusion}

We study auditing of differential privacy in the black-box setting, empirical auditing image classification models when trained with DP-SGD.  Focusing on one-run auditing methods, we make the observation that quantile-regression based MIA approaches complement the computationally efficient nature of one-run procedures introduced by \citet{steinke2023privacy} and \citet{mahloujifar2024auditing}. Empirically, we demonstrate that our quantile-regression based approach improves the baseline procedures across a wide range of data settings. We recognize, however, that our experiments show that large gap between empirical and theoretical privacy still exists. We hope that our work will help inspire future studies that may attempt to further close this gap in the black-box auditing setting.

\bibliography{docs/main}
\bibliographystyle{tmlr}

\end{document}